\documentclass[graybox]{svmult}

\usepackage{type1cm}        
\usepackage{makeidx}         
\usepackage{graphicx}        
\usepackage{multicol}        
\usepackage[bottom]{footmisc}

\usepackage{newtxtext}       %
\usepackage[varvw]{newtxmath}       

\usepackage[ruled,vlined]{algorithm2e} 
\usepackage{booktabs} 
\usepackage{multirow} 

\makeindex             

\begin{document}

\title*{LLM-Based Feature Extraction for Multi-Criteria Inventory Classification}
\author{
Humza Akhtar\orcidID{0000-0003-1594-2204} 
\and 
Rami Pinto\orcidID{0009-0000-4520-3133}
}
\institute{
MongoDB Inc., New York, NY, USA\\
\email{humza.akhtar@mongodb.com}, 
\email{rami.pinto@mongodb.com}
}
%
%
\maketitle

\abstract*{E-commerce supply chains generate large volumes of customer feedback and operational data that remain underutilized in inventory management. Traditional inventory classification techniques such as ABC analysis rely primarily on annual dollar usage, while multi-criteria inventory classification (MCIC) methods extend this but still depend on structured quantitative data and do not utilize unstructured information available in modern supply chains. This paper proposes an LLM-based feature extraction approach that transforms unstructured sources such as customer reviews into quantitative features using vector embeddings and LLM-generated evaluation criteria. The enriched feature set is incorporated into an MCIC framework based on the Hadi-Vencheh model. Experimental results show measurable reclassification effects when customer sentiment signals are included, demonstrating that integrating unstructured data into MCIC can enhance inventory prioritization.}

\abstract{E-commerce supply chains generate large volumes of customer feedback and operational data that remain underutilized in inventory management. Traditional inventory classification techniques such as ABC analysis rely primarily on annual dollar usage, while multi-criteria inventory classification (MCIC) methods extend this but still depend on structured quantitative data and do not utilize unstructured information available in modern supply chains. This paper proposes an LLM-based feature extraction approach that transforms unstructured sources such as customer reviews into quantitative features using vector embeddings and LLM-generated evaluation criteria. The enriched feature set is incorporated into an MCIC framework based on the Hadi-Vencheh model. Experimental results show measurable reclassification effects when customer sentiment signals are included, demonstrating that integrating unstructured data into MCIC can enhance inventory prioritization.}


\section{Introduction}
\label{intro}

Inventory classification is a central task in supply chain management, particularly in e-commerce contexts where large product catalogs must be prioritized under uncertain demand, variable delivery performance, and high customer sensitivity to service quality. Traditional ABC analysis remains widely used because of its simplicity and interpretability. It classifies inventory items into categories A, B, and C according to annual dollar usage, thereby allowing firms to concentrate resources on high-value items \cite{partovi2002classifying,chu2008controlling}. However, reliance on annual dollar usage alone can overlook operational and market-related signals such as lead time, commonality, criticality, durability, and customer-perceived service quality.

Multi-criteria inventory classification (MCIC) addresses this limitation by incorporating additional attributes into the classification process. Prior work has proposed several approaches, including cross-tabulation, clustering, analytic hierarchy process, genetic algorithms, and artificial-intelligence-based classification methods \cite{flores1987implementing,ernst1990operations,partovi1993using,guvenir1998multicriteria,yu2011multi}. These methods improve upon traditional ABC analysis by considering multiple structured attributes. Nevertheless, most MCIC approaches still depend on quantitative data that are already available in structured enterprise systems.

At the same time, modern supply chains generate large volumes of unstructured data, including customer reviews, supplier reports, service notes, quality records, messages, and, increasingly, multimodal evidence such as images or videos. Such data may contain signals about customer satisfaction, delivery reliability, product quality, and perceived risk. Recent advances in LLMs and embedding models provide a practical way to 
transform unstructured evidence into machine-readable representations 
\cite{dubey2024llama3}. This creates an opportunity to extend MCIC models with qualitative features derived from unstructured data.

This paper contributes an LLM-based feature extraction framework for AI-augmented MCIC, integrating LLM-derived features into the Hadi-Vencheh model \cite{hadi2010improvement} and evaluating the approach on a public e-commerce dataset with customer reviews, order, and delivery data.

\section{Related Work}
\label{litreview}

The limitations of single-criterion ABC analysis have been widely 
discussed. Flores and Whybark \cite{flores1987implementing} introduced 
a bi-criteria cross-tabulation approach, while subsequent work applied 
clustering \cite{ernst1990operations}, the analytic hierarchy process 
\cite{partovi1993using}, and genetic algorithms \cite{guvenir1998multicriteria}. Ng \cite{ng2007inventory} proposed a simple weighted classifier, later improved by Hadi-Vencheh \cite{hadi2010improvement} to reduce reliance on exact subjective weights. Yu \cite{yu2011multi} demonstrated that machine learning techniques including support vector machines and \textit{k}-nearest neighbors can outperform traditional statistical methods in MCIC. Recent reviews and extensions confirm that MCIC continues to benefit from machine learning advances \cite{douissa2020non,khanorkar2023selective}.

Despite these developments, existing MCIC methods primarily operate on structured quantitative attributes. Recent work has extended these methods with additional machine learning techniques in manufacturing contexts \cite{albayrak2025multicriteria}, confirming that structured criteria continue to dominate the field. Separately, LLMs have been applied to extract and classify entities from unstructured supply chain text \cite{liu2024supplychain}, demonstrating the feasibility of LLM-based feature extraction in operational settings. However, to the authors' knowledge, no prior work has integrated LLM-derived features from unstructured data into established MCIC frameworks, representing the gap this paper addresses.

\section{Methodology}
\label{methodology}

\begin{figure}[htbp]
\centering
\vspace{-4pt}
\includegraphics[width=0.92\linewidth]{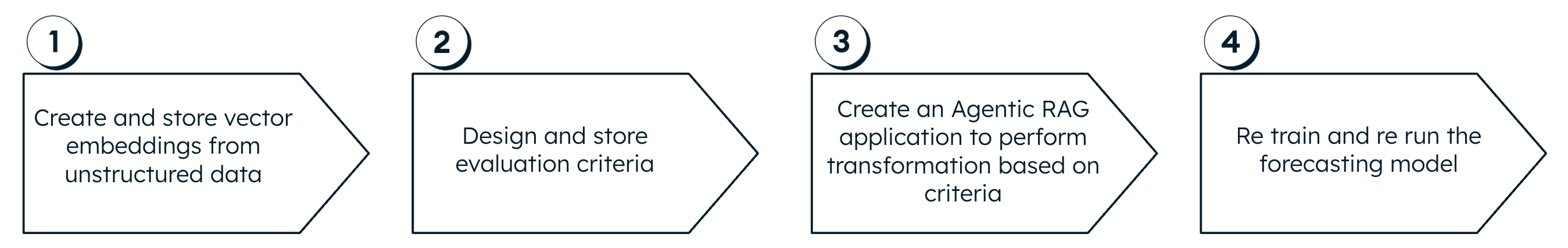}
\vspace{-8pt}
\caption{Four-stage framework for LLM-based feature extraction and 
AI-augmented inventory classification}
\vspace{-4pt}
\label{fig:methodology}
\Description{A four-stage methodology diagram showing: (1) Create and 
store vector embeddings from unstructured data, (2) Design and store 
evaluation criteria, (3) Create an application to perform transformation 
based on criteria, and (4) Re-train and re-run the forecasting model.}
\end{figure}

The proposed framework extends MCIC by adding LLM-derived qualitative features to conventional structured inventory attributes. The process consists of four stages: collecting item-level data, defining qualitative evaluation criteria, transforming unstructured evidence into numeric feature scores, and rerunning the MCIC model with the enriched feature set. The framework is implemented using MongoDB as the data 
platform. Structured inventory attributes and unstructured text sources 
are stored as documents in MongoDB collections, while vector embeddings 
generated from textual data are stored and queried using MongoDB 
Vector Search. For each product and criterion, semantically relevant 
evidence is retrieved via approximate nearest neighbor search over the 
embedding space, enabling efficient retrieval at scale without requiring 
exact keyword matches. 

Let $P_r$ denote the set of inventory items, $D_s$ the structured data sources, and $D_u$ the unstructured data sources. Structured attributes may include annual dollar usage, average unit cost, and lead time. Unstructured sources may include reviews, service notes, or inspection images, represented through suitable embeddings for semantic retrieval.

A qualitative criterion $c \in C$ is defined by a name, a textual definition, one or more evaluation elements, and a scoring scale. Domain experts provide the business objective and validate the resulting criterion. An LLM-assisted process translates the objective into an operational scoring rule. For example, a perceived delivery experience criterion may assign higher scores to items associated with negative delivery sentiment, since such items may require tighter monitoring or inventory control.

For each product $p \in P_r$ and criterion $c \in C$, relevant evidence is retrieved from $D_u$. The retrieved context is scored according to the criterion definition using an LLM-assisted evaluation process, producing an element score $s_e$. If a criterion contains multiple elements, the element scores are aggregated as

\begin{equation}
s_c(p)=\sum_{e \in c} w_e s_e
\end{equation}

where $w_e$ denotes the relative importance of element $e$. The resulting score $s_c(p)$ is added to the structured feature vector of product $p$. All criteria are oriented such that larger values indicate a stronger need for managerial attention before normalization.

The final classification step follows the Hadi-Vencheh model \cite{hadi2010improvement}, an extension of the Ng model \cite{ng2007inventory}. Let $i=1,\dots,n$ denote inventory items and $j=1,\dots,J$ denote the selected criteria. The value of item $i$ under criterion $j$ is denoted by $y_{ij}$. Each criterion is normalized to the interval $[0,1]$:

\begin{equation}
s_{ij} =
\frac{y_{ij}-\min(y_j)}
{\max(y_j)-\min(y_j)}
\end{equation}

The overall score for each item is calculated using a weighted optimization model. The weights are constrained by an expert-defined ranking of criteria rather than fixed exact values:

\begin{equation} \begin{aligned} \max \quad & S_i = \sum_{j=1}^{J} s_{ij} w_j \\ \text{s.t.} \quad & \sum_{j=1}^{J} w_j^2 = 1 \\ & w_j \geq w_{j+1} \geq 0, \quad j = 1 \dots, J-1 \\ & w_j \geq 0, \quad j = 1, \dots, J \end{aligned} \label{eq:hadi-vencheh} \end{equation}

The model produces a composite score $S_i$ for each item. Items are then ranked and assigned to classes A, B, or C. Traditional ABC analysis is used as the baseline classification, while conventional MCIC and AI-augmented MCIC are compared to assess the impact of adding LLM-derived features.

\section{Experimental Evaluation}
\label{discussion}

The empirical evaluation uses the Brazilian E-Commerce Public Dataset by Olist \cite{olist_andr__sionek_2018}, containing over 100,000 anonymized orders with product, order, delivery, and review information.


Three classification scenarios are evaluated. First, traditional ABC analysis is performed using annual dollar usage as the sole criterion. This produces a baseline distribution of 16.0\% Class A items, 28.6\% Class B items, and 55.4\% Class C items, consistent with the Pareto principle. Second, a conventional MCIC model is applied using annual dollar usage (ADU), average unit cost (AUC), and lead time (LT). Third, an AI-augmented MCIC model (AMCIC) is evaluated by adding a perceived delivery experience (PDE) feature extracted from customer reviews. 

The PDE feature is operationalized as a scalar criterion derived from review retrieval and criterion-based scoring. It complements objective lead time by capturing the subjective customer impact of delivery performance. Negative delivery experiences receive higher scores because they indicate greater potential need for managerial attention, while consistently positive experiences receive lower scores.

To quantify the effect of reclassification, transition matrices were generated comparing class changes between traditional ABC and MCIC, between MCIC and AMCIC, and between traditional ABC and AMCIC. The matrices indicate whether items are retained in the same class, moved to a higher-priority class, or moved to a lower-priority class.

\begin{figure}[htbp]
\centering
\vspace{-4pt}
\includegraphics[width=\linewidth]{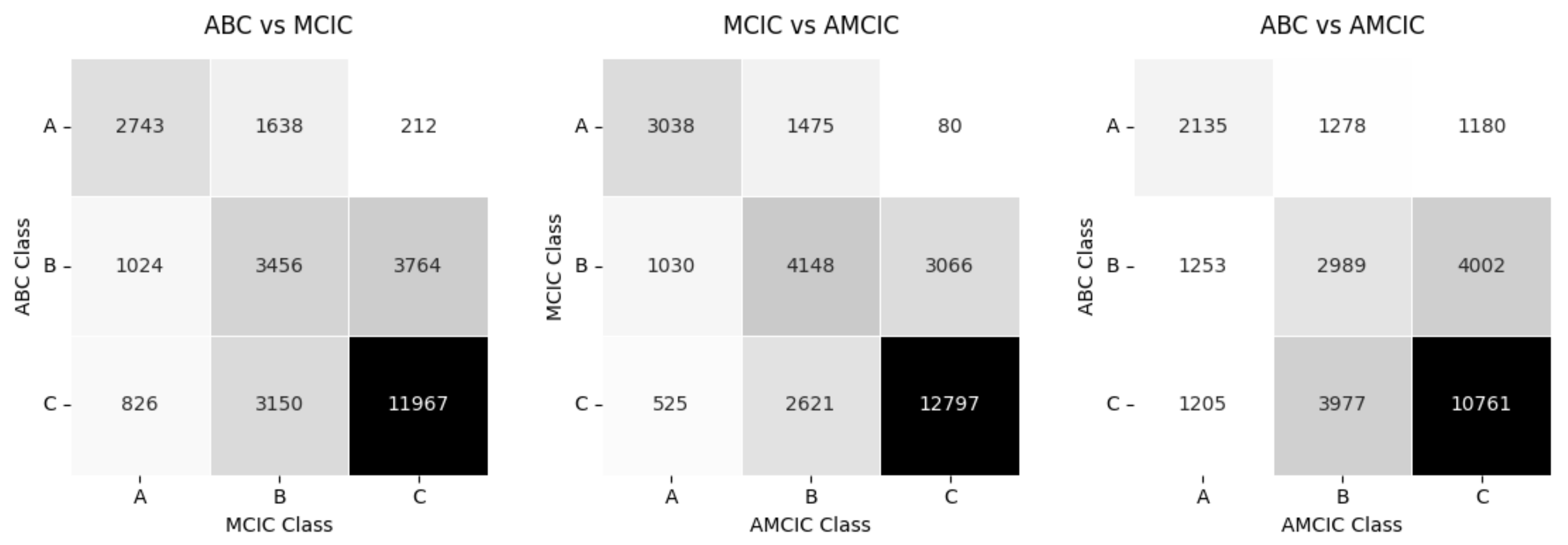}
\caption{Transition matrices showing inventory class changes across methods}
\label{fig}
\Description{Three transition matrices comparing inventory class changes between traditional ABC and MCIC, MCIC and AI-augmented MCIC, and traditional ABC and AI-augmented MCIC. Each matrix shows how many items remain in the same class or move between classes A, B, and C.}
\end{figure}
\vspace{-4pt}

A consolidated comparison of the class transitions is provided in Table~\ref{table}. Moving from traditional ABC to MCIC results in a total retention rate of 63.12\%, with 17.37\% of items moving to a higher-priority class and 19.51\% moving to a lower-priority class. Moving from MCIC to AMCIC results in a higher retention rate of 69.43\%, with 14.51\% of items moving upward and 16.06\% moving downward. The direct comparison between ABC and AMCIC shows a retention rate of 55.19\%, indicating that the combined effect of structured MCIC attributes and the LLM-derived PDE feature substantially changes the prioritization of some items.

\begin{table}[!t]
\caption{Class transition summary showing retention, increase, and decrease rates}
\label{table}
\begin{tabular*}{\textwidth}{@{\extracolsep{\fill}}lccccccccc@{}}
\toprule

\multirow{2}{*}{Class}
& \multicolumn{3}{c}{ABC $\rightarrow$ MCIC}
& \multicolumn{3}{c}{MCIC $\rightarrow$ AMCIC}
& \multicolumn{3}{c}{ABC $\rightarrow$ AMCIC} \\

\cmidrule(lr){2-4}
\cmidrule(lr){5-7}
\cmidrule(lr){8-10}

& Ret. & Inc. & Dec.
& Ret. & Inc. & Dec.
& Ret. & Inc. & Dec. \\

\midrule

A
& 59.72 & 0.00 & 40.28
& 66.14 & 0.00 & 33.86
& 46.48 & 0.00 & 53.52 \\

B
& 41.92 & 12.42 & 45.66
& 50.32 & 12.49 & 37.19
& 36.26 & 15.20 & 48.54 \\

C
& 75.06 & 24.94 & 0.00
& 80.27 & 19.73 & 0.00
& 67.50 & 32.50 & 0.00 \\

\midrule

Total
& 63.12 & 17.37 & 19.51
& 69.43 & 14.51 & 16.06
& 55.19 & 22.36 & 22.45 \\

\bottomrule
\end{tabular*}
\end{table}

The results show that LLM-derived features provide interpretable refinements to inventory classification. Items promoted under AMCIC may represent cases where structured metrics understate service risk, while downward movements may indicate low perceived delivery risk despite strong quantitative profiles. The approach therefore introduces an additional signal for differentiated monitoring, replenishment review, and managerial prioritization. These results demonstrate reclassification effects rather than validated improvements in downstream operational performance; whether shifts translate to better service levels or lower holding costs requires validation against operational ground truth and is a direction for future work.

\section{Conclusion and Future Research}
\label{conclusion}

This paper presented an LLM-based feature extraction framework for enriching multi-criteria inventory classification with information derived from unstructured operational data. We identified three limitations of existing approaches: the reliance of traditional ABC analysis on annual dollar usage as a sole criterion, the dependence of conventional MCIC methods on structured quantitative attributes, and the underutilization of unstructured data available in modern supply chains. The proposed framework addresses all three. It moves beyond single-criterion ABC analysis by incorporating multiple criteria into a weighted classification model. It extends conventional MCIC by introducing LLM-derived qualitative features alongside structured attributes and it operationalizes unstructured customer feedback as a quantitative inventory signal, demonstrating that such data can meaningfully influence item  prioritization. 

Further work should validate whether reclassification translates into measurable operational gains by testing against datasets with observable ground-truth outcomes such as service level, stockout rate, and holding cost. A natural extension is to couple classification with demand forecasting models, where LLM-derived sentiment features may improve accuracy for high-risk items. Robustness checks and bias detection for LLM-generated scores remain important requirements for practical deployment.

\end{document}